\documentclass{article}
\usepackage[bottom]{footmisc}
\usepackage{float}
\usepackage{sidecap}
\usepackage{multirow}
\usepackage{parskip}

\usepackage{natbib}
\setcitestyle{authoryear,open={(},close={)}} 

\usepackage[margin=1.25in]{geometry}
\usepackage[utf8]{inputenc} 
\usepackage[T1]{fontenc}    
\usepackage{url}            
\usepackage{booktabs}       
\usepackage{amsfonts}       
\usepackage{nicefrac}       
\usepackage{microtype}      
\usepackage{kpfonts}
\usepackage{graphicx}
\usepackage{multicol}
\usepackage{enumitem}

\usepackage[small]{caption}
\usepackage[table]{xcolor}
\definecolor{codegreen}{rgb}{0,0.6,0}
\definecolor{codegray}{rgb}{0.3607843137,
0.4823529412,
0.5725490196}
\definecolor{codeblue}{rgb}{0,0.28,0.67}
\definecolor{backcolour}{rgb}{0.9882352941,
0.9725490196,
0.9294117647}

\usepackage{hyperref}
\hypersetup{
    colorlinks,
    linkcolor={codeblue},
    citecolor={codeblue},
    urlcolor={codeblue}
}

\usepackage[scaled]{beramono}

\usepackage{lipsum}
\usepackage{listings}

\lstdefinestyle{mystyle}{
    escapechar=\%,
    backgroundcolor=\color{backcolour},   
    commentstyle=\color{codegray},
    keywordstyle=\color{codeblue},
    basicstyle=\ttfamily\small,
    breakatwhitespace=false,         
    breaklines=true,                 
    captionpos=b,                    
    keepspaces=true,                 
    numbers=left,                    
    numbersep=5pt,                  
    showspaces=false,                
    showstringspaces=false,
    showtabs=false,                  
    tabsize=2
}
\usepackage{amsthm}

\usepackage{setspace}

\begin{document}

\thispagestyle{empty}

\renewcommand*{\thefootnote}{\fnsymbol{footnote}}
\begin{center}
\Large{\textsc{ARWKV: Pretrain is not what we need, an RNN-Attention-Based Language Model Born from Transformer}}
\vspace{0.8cm}
\normalsize 
\textbf{\textbf{\\ Lin Yueyu \& Li Zhiyuan \& Peter Yue \& Liu Xiao \\}}
\vspace{0.1cm}
\textit{yueyu.lin@me.com \\
lizhiyuan@uniartisan.com \\
ziyinyue07@gmail.com \\
liu.xiao.in@gmail.com \\}

\medskip
\normalsize
\end{center}

\vspace{0.4cm}

\begin{abstract}
As is known, hybrid quadratic and subquadratic attention models in multi-head architectures have surpassed both Transformer and Linear RNN models \cite{dong2024hymba}, with these works primarily focusing on reducing KV complexity and improving efficiency. For further research on expressiveness\footnote{https://github.com/Jellyfish042/Sudoku-RWKV, https://github.com/Jellyfish042/RWKV\_Othello}, we introduce our series of models distilled from Qwen 2.5, based on pure native RWKV-7 attention, which aims to make RNN more expressive and demonstrates state tracking ability beyond transformers. We work with QRWK 32B \footnote{https://huggingface.co/recursal/QRWKV6-32B-Instruct-Preview-v0.1} based on RWKV-6 architecture, another approach that reduces the entire knowledge processing time to just 8 hours using 16 AMD MI300X GPUs while maintaining  Qwen 2.5's performance. In fact, the distillation process can utilize any LLM, not just Qwen, and enables knowledge transfer from larger LLMs to smaller ones with more fewer tokens. We will explain the detailed process and share our insights on building more powerful foundation models. Please note that this is an ongoing work that will be updated continuously. The model checkpoints and source code are available at \href{https://github.com/yynil/RWKVInside}{https://github.com/yynil/RWKVInside}, \href{https://huggingface.co/RWKV-Red-Team/ARWKV-7B-Preview-0.1}{https://huggingface.co/RWKV-Red-Team/ARWKV-7B-Preview-0.1}.    
\end{abstract}

\medskip

\tableofcontents
\section{Introduction}
The emergence of Linear RNNs (LRNNs) has grown rapidly, with models like RWKV \cite{peng2024eagle}, DeltaNet \cite{yang2024parallelizing}, Mamba-2 \cite {wang2024mamba}, GLA \cite{yang2023gated}, and others showing strong competitiveness with transformers. However, these models face natural penalties in context learning and long-context retrivel due to subquadratic attention limitations. Recently, RWKV-7 introduced a promising architecture, where a 0.1B parameter model achieved perfect results in 16k passkey retrieval\footnote{https://x.com/BlinkDL\_AI/status/1869433254425833487}. With its transition matrix having wider eigenvalues \cite{grazzi2024unlocking}, it demonstrates stronger state tracking capabilities than transformers \cite{merrill2024illusion}.
While Qwen 2.5 was trained on 18 trillion tokens requiring enormous GPU resources, making pretraining impractical for academic use, we bridge this gap by refining the distillation method. Our approach enables training a 7B parameter model on a single A100 80G GPU, while 4x8 A100s can handle a 32B model.
\bigskip

\section{Architecture}
We know qwen 2.5 dense models is the Transformer-based decoder architecture with group query attention, SwiGLU activation
 function for non-linear activation, Rotary Positional Embeddings for encoding  position information, QKV bias in the attention mechanism and RMSNorm \cite{yang2024qwen2}. We only keep RMSnorm and SwiGLU activiation and replace the rest with rwkv-7 attention.In GQA, queries are grouped while keys and values remain separate:
 
\begin{align}
\text{GQA}(Q_g, K, V) = \text{concat}\left[\text{head}_1, ..., \text{head}_g\right]W^O     
\end{align}

where each head is computed as:
\begin{align}
\text{head}_i = \text{Attention}(Q_iW^Q_i, KW^K_i, VW^V_i)  
\end{align}

In RWKV-7 attention which is the time mixing:

\begin{center}
\begin{align}
    \textbf{State}_t = \textbf{State}_{t-1} \left({diag}(w_{t}) - \hat{\kappa}^T_t (a_t \odot \hat{\kappa}_t)\right) + v^T_t \cdot \tilde{k}_t 
\end{align}
\end{center}
the state means matrix-valued attention state.where a is the in-context learning rate.

In the comparing with RWKV-6 \cite{peng2024eagle}:
\begin{center}
\begin{align}
 state' = \text{diag}(w) \cdot state + k^\top \cdot v   
\end{align}
\end{center}

then we just relpace self-attention in every layer with RWKV-7 time mixing module in Figure 1.

\begin{figure}[h]
    \centering
    \includegraphics[width=0.7\linewidth]{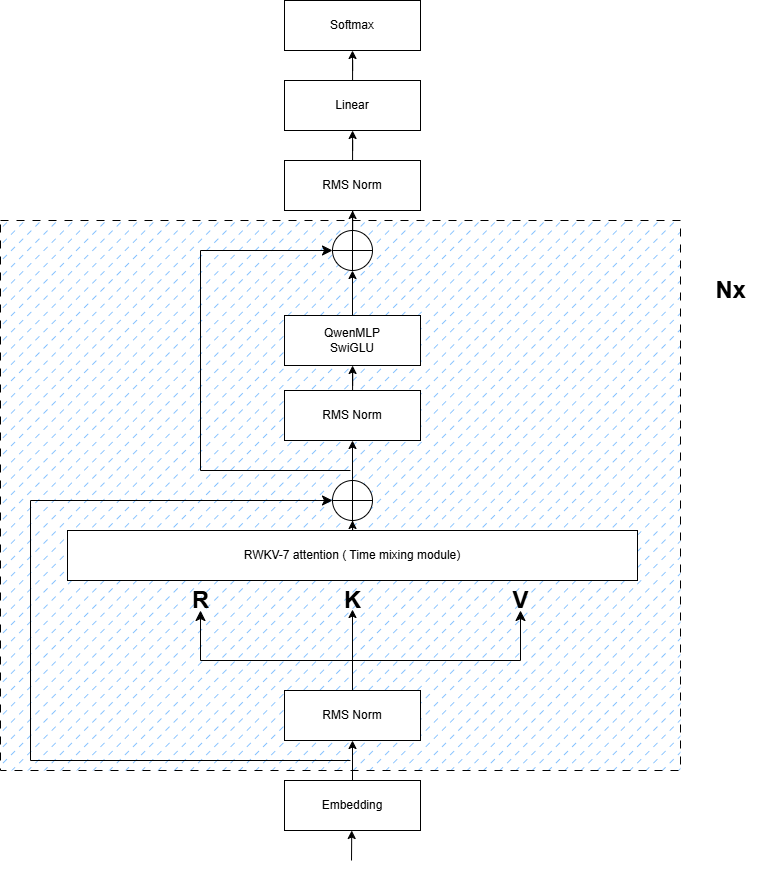}
    \caption{replace self-attention by RWKV-7 time mixing module}
    \label{fig:enter-label}
\end{figure}

\bigskip

\section{From Transformer to RNN}
Inspired by MambaInLLaMA \cite{wang2024mamba} and Phi-Mamba, we aimed to produce a pure RNN model similar to QRWKV. We tested both layer-wise distillation and one-step distillation methods, finding the latter to be more efficient. Our research revealed that using attention alignment in stage 1 is crucial for maintaining the original model's performance. By mimicking self-attention through RWKV's time mixing module, we can preserve attention expressiveness while transforming the state attention to function more like a meta-learner \cite{hospedales2021meta}.

\bigskip

\subsection{Stage 1 - Time Mixing module replacing Self-Attention}
In this stage,we align the hidden state output between the student and teacher attention block \cite{bick2024transformers}, frozen the MLP and remove the group normalization for the attention output and make the gate initialized with 1 .[Figure \ref{stage-1}] Context length can dramaticaly increase the trainning time, in this work we train 2048 with one h800

\begin{align}
L_{special} = \left\|\mathbf{h}_{teacher} - \mathbf{h}_{student}\right\|_2 \cdot (d_{model})^{-0.5}
\end{align}
where $d_{model}$ is dimension size of hidden state.

We found that initializing state attention from teacher's attention is not necessary.
 The convergence speed and final values of the loss [Figure \ref{fig:stage-1-loss}] indicate the target attention ability to capture the teacher model's  internal attention representations. \cite{bick2024transformers}

 due to the fixed state size in rwkv-7 time mixing module, we can see it as a compression process \cite{skean2024does} ,  or it can be seen attention as a map between probability measures \cite{castin2024smooth}, 

\subsection{Stage 2 - Knowledge Distillation}
Divergence-based methods minimize diver
gence between the probability distributions of the teacher
 and student models \cite{xu2024survey}, We adopt word-level KL-Divergence instead of sequence-level knowledge distillation (SeqKD).In practice, we distill from 32B to 7B model. 
 
 For the dataset, we balance its distribution \cite{xu2024magpie} based on stage 1 training, achieving fast convergence with only 20M tokens.

We then experiment with whether or not to freeze the MLP layer and introduce a gate-free technique which disables the gate mechanism entirely

\subsection{Stage 3 - SFT and DPO}
During this stage, we use supervised fine-tuning (SFT) to expand the model's context length and Direct Preference Optimization (DPO) to align with user preferences. Our training process involves 20M tokens in stage 1, 40M tokens in stage 2, and 770M tokens in stage 3 for context length extension.

\section{Evaluation}
We conducted ablation experiments between 7B models distilled using different approaches in stage 2. By controlling the presence of Gate, freeze MLP, size of teacher model, we distill model not-freezing MLP(ARWKV-M), model with GATE and not freezing MLP(ARWKV-G-M), model without Gate nor freezing MLP(ARWKV) and model with teacher model as QWEN2.5-32B-Instruct(ARWKV-from32B). All models are distilled from QWEN2.5-7B-Instruct except ARWKV-from32B. After stage-2 training, all models are tested with several benchmarks to demonstrate the impact of different training factors on the final model performance.

Moreover, we found that although we trained the model using bfloat16 (BF16), performing inference with float16 (FP16) significantly improved the performance, This differs from the original RWKV implementation, which required careful tuning of the layer scaling parameters to avoid overflow.

Our analysis of Table 1 reveals that knowledge distillation from the 32B parameter model, when performed without gating mechanisms and with frozen MLPs, yields suboptimal results. This performance degradation may be attributed to the limited capacity of the 7B model's MLP layers to effectively accommodate and adapt to the more sophisticated attention patterns learned by the 32B model's architecture. This observation suggests a potential architectural mismatch in the direct transfer of attention mechanisms between models of significantly different scales.

\begin{table}[H]
    \centering
    \renewcommand{\arraystretch}{1.5}
\begin{tabular}{l|l|l|l|l|l}
\toprule
 & Qwen2.5-7B-Instruct & ARWKV& active MLP & w/ gate \& active MLP & ARWKV-from32B \\
\hline
\midrule
MMLU           & 71.72  & 62.41  & 58.22 & 64.77 & 61.78 \\
\hline
Squad      & 47.89  & 40.05  & 40.35 & 38.74 & 39.01 \\
\hline
GPQA(Diamond)           & 49.0  & 45.5  & 51.1 &  &  \\
\hline
WinoGrande           & 71.35  & 68.67  & 69.67 & 68.98 & 68.35 \\
\hline
GSM8K          & 82.34  & 39.95  & 51.93 & 47.99 & 43.44 \\
\hline
IfEval       & 73.62  & 52.16  & 48.68 & 52.16 & 44.12 \\
\hline
Arc-c       & 54.86  & 52.22  & 53.52 & 52.22 & 50.77 \\
\hline

\end{tabular}
    \caption{This is a ongoing work, benchmark based on stage-2 , currently we limit the context length to 2048 and use same datasets}
    \label{tab:my_label}
\end{table}

\bigskip

\section{Conclusions}

\bigskip

We demonstrate that attention alignment combined with knowledge distillation can effectively transform transformer attention patterns into RNN-based attention in a straightforward manner.In some hybrid architecture ,this attention module serves as a memory component \cite{dong2024hymba}, different kind of attention introduce unique inductive bias \footnote{https://medium.com/@felixhill/the-agreeable-lesson-9766382c6d83} , bring more expressive power. 

\section{Future Work}

For our subsequent phase of investigation, we will implement Stage 3 post-training to replicate the reasoning capabilities demonstrated by deepseek-R1 \cite{guo2025deepseek} models.

Furthermore, we propose to generalize this methodology across diverse architectural paradigms, encompassing Mixture-of-Experts (MoE) frameworks, multimodal architectures, hybrid architectures and model compression scenarios. This expansion aims to validate the robustness and transferability of our approach across different computational paradigms.

\section{Acknowledgements}
\bigskip

We would like to express our gratitude to Yuhang He for his endorsement.

\bibliography{references}

\clearpage

\begin{figure}[h]
    \centering
    \includegraphics[width=1\linewidth]{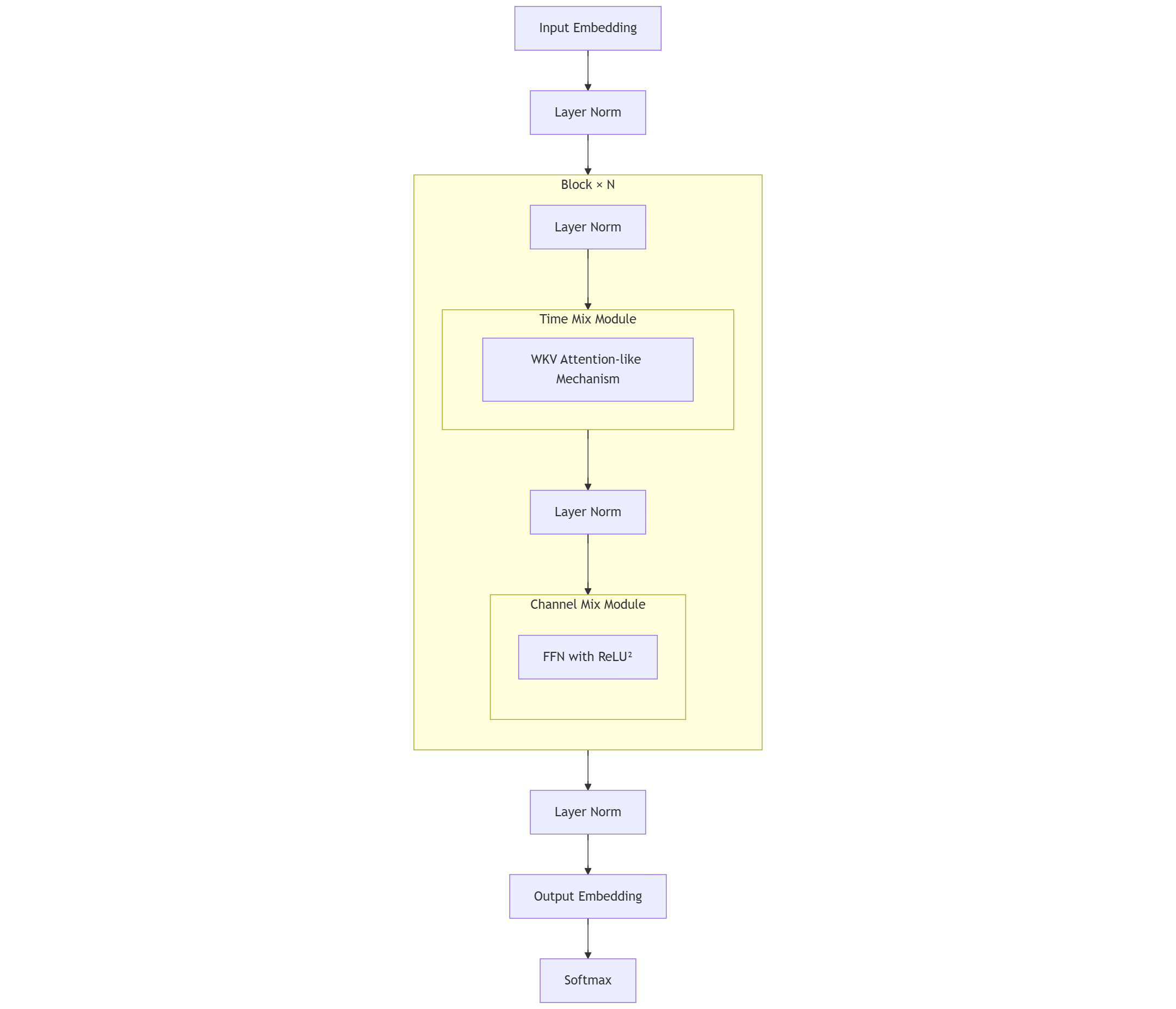}
    \caption{RWKV-7 architecture.capability of attention is the key for RNN-based LLMs, which in this case is Time mixing module}
    \label{fig:rwvk-7-arch}
\end{figure}

\begin{figure}[h]
    \centering
    \includegraphics[width=0.3\linewidth]{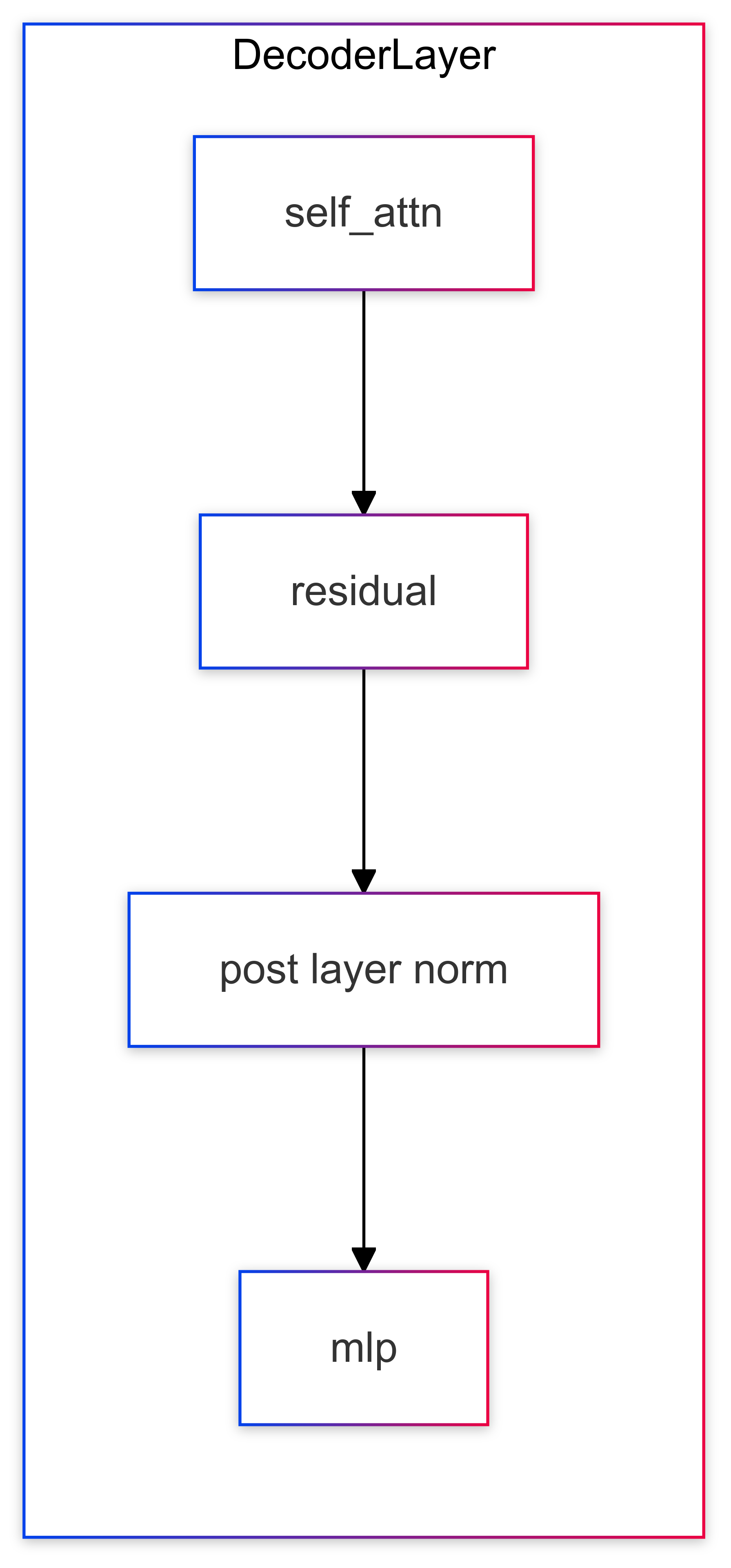}
    \caption{General Decoder Layer in transformer}
    \label{fig:g_decoder}
\end{figure}

\begin{figure}[h]
    \centering
    \includegraphics[width=0.5\linewidth]{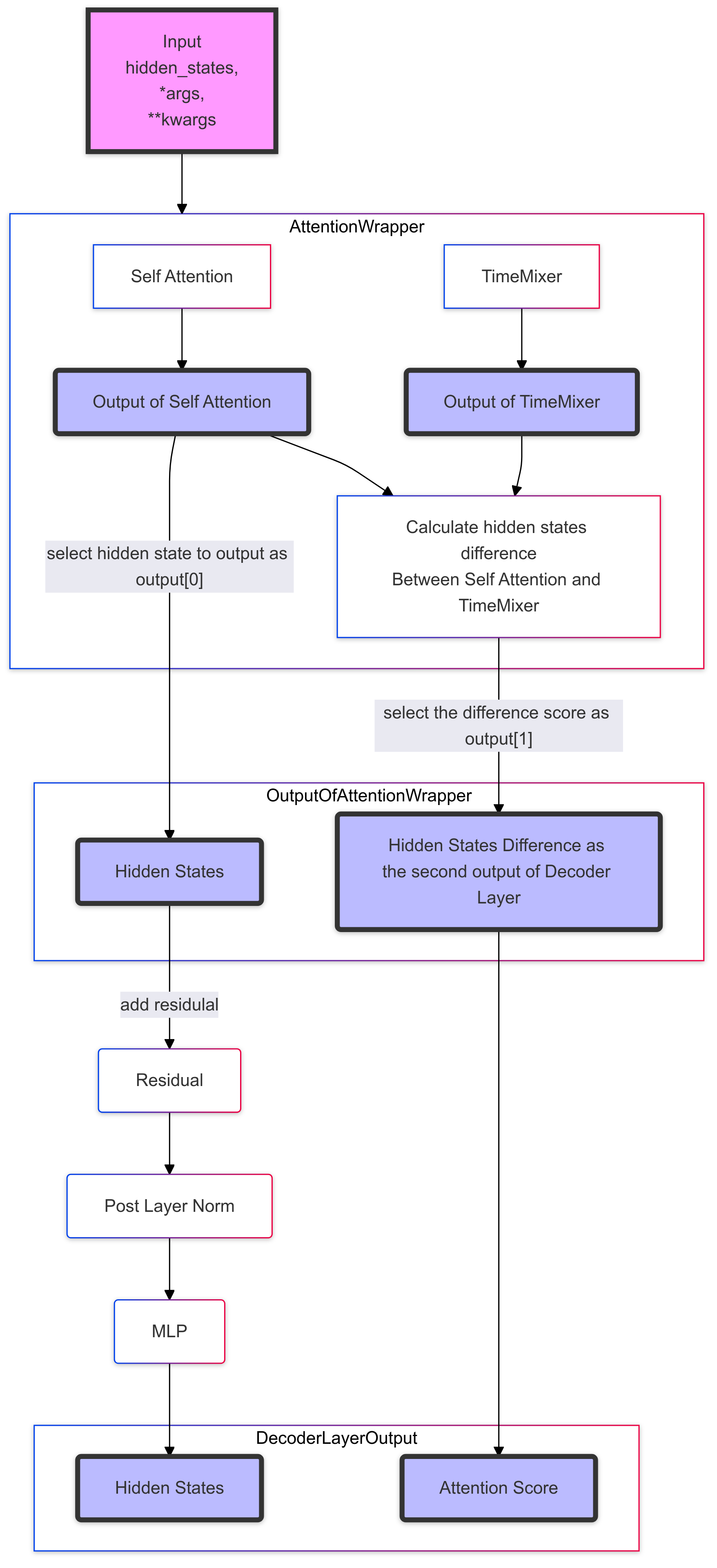}
    \caption{We replace the standard Attention with an AttentionWrapper that contains both the original self-attention mechanism and a TimeMixer. The TimeMixer is trained to minimize the gap between its output and that of the self-attention module. The final output combines the hidden states from the original self-attention with the residual difference between self-attention and TimeMixer outputs. This architecture enables the model to optimize the TimeMixer to progressively reduce the discrepancy between self-attention and TimeMixer outputs.}
    \label{stage-1}
\end{figure}

\begin{figure}[h]
    \centering
    \includegraphics[width=1\linewidth]{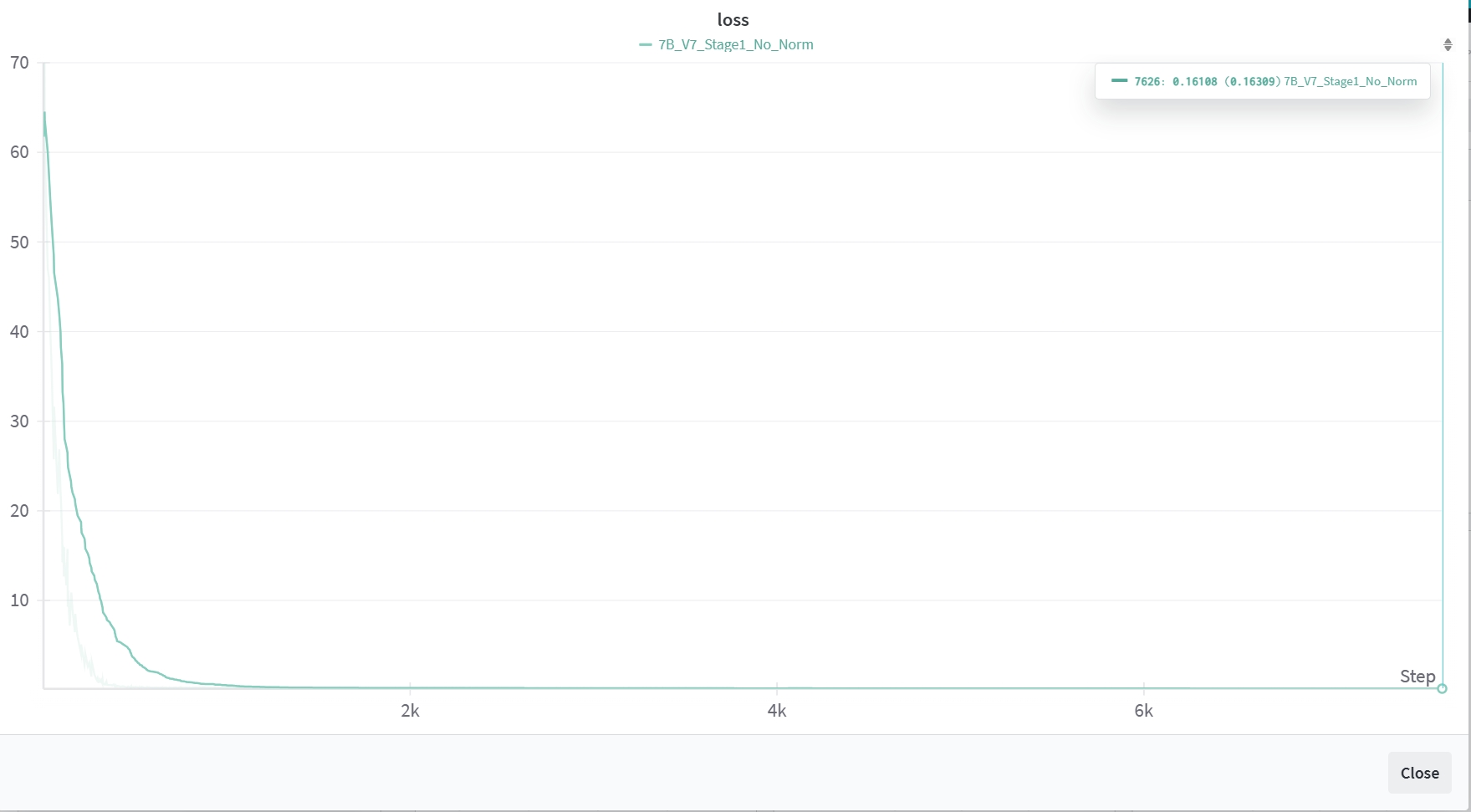}
    \caption{Stage-1 loss, 18 hours with one 8*h800 80G , context length 2048, 4B tokens}
    \label{fig:stage-1-loss}
\end{figure}

\end{document}